\documentclass{article} % For LaTeX2e
\usepackage{iclr2027_conference,times}

\usepackage{amsmath,amsfonts,bm}

\def\eqref#1{equation~\ref{#1}}
\def\1{\bm{1}}

\DeclareMathAlphabet{\mathsfit}{\encodingdefault}{\sfdefault}{m}{sl}
\SetMathAlphabet{\mathsfit}{bold}{\encodingdefault}{\sfdefault}{bx}{n}

\usepackage{hyperref}
\usepackage{url}

\usepackage{lipsum}
\usepackage{graphicx}
\usepackage{subcaption}
\usepackage{booktabs}
\usepackage{multirow}
\usepackage[table]{xcolor}
\usepackage{float}
\usepackage{amsmath}
\usepackage{wrapfig}
\definecolor{rowhl}{gray}{0.92}
\newcommand{\hlc}{\cellcolor{rowhl}}
\newcommand{\system}{VLA-ZO}

\title{\system: Fast Zeroth-Order Adaptation\\for Vision-Language-Action Models}

\author{Jaemin Kim, Jiahn Kim, Taesik Gong\thanks{Corresponding author.} \\
UNIST \\
\texttt{\{jm611,amssoez112,taesik.gong\}@unist.ac.kr}
}

\iclrfinalcopy % Uncomment for camera-ready version, but NOT for submission.
\begin{document}

\setcounter{footnote}{1}
\maketitle
\lhead{}
\renewcommand{\headrulewidth}{0pt}

\begin{abstract}
Adapting vision-language-action (VLA) models to deployment-time distribution shifts is important for reliable robotic operation, but conventional first-order adaptation can exceed the memory budget of inference-oriented deployment platforms. Zeroth-order (ZO) optimization offers a forward-only alternative with inference-level memory, but accurate gradient estimation requires many perturbation queries, making naive ZO prohibitively slow for large VLA models. We present VLA-ZO, a framework for fast ZO adaptation that exploits the structure of VLA computation. By confining adaptation to the action side, VLA-ZO keeps the expensive vision-language prefix frozen and reuses its conditioning states across perturbation queries and optimizer steps, while schedule-aware prefetching hides state-transfer overhead. On LIBERO camera-viewpoint shifts, VLA-ZO reduces end-to-end adaptation time by 25.59$\times$ at $q=16$ and 32.54$\times$ at $q=64$ relative to baseline ZO, while improving average task success from 48.27\% without adaptation to 58.17\% and 63.58\%, respectively. These results show that making ZO faster can make larger query budgets practical, providing a promising path toward resource-efficient VLA adaptation on deployment platforms.
\end{abstract}

\section{Introduction}
Vision-Language-Action (VLA) models are emerging as a general interface for robotic control~\citep{kim2024openvla, BlackK-RSS-25, black2025pi}, but deploying a fixed policy is often insufficient in real-world environments.
Changes in camera viewpoint, appearance, and operating conditions shift the observation distribution away from that seen during training, causing VLA policies to degrade sharply under such shifts.
Specifically, the success rates of state-of-the-art VLAs collapse from $95\%$ to below $30\%$ under modest perturbations to camera pose and robot initial state~\citep{fei2025liberoplusindepthrobustnessanalysis}.
Recent work has explored one-shot adaptation~\citep{Li_2026_CVPR, 10.1007/978-3-032-37574-2_24}, in which a VLA policy is adapted using a single demonstration from a target environment.
This setting enables data-efficient post-deployment adaptation to address these distribution shifts.

\begin{figure}[t]
  \centering
  \begin{subfigure}[b]{0.66\linewidth}
    \centering
    \includegraphics[width=\linewidth]{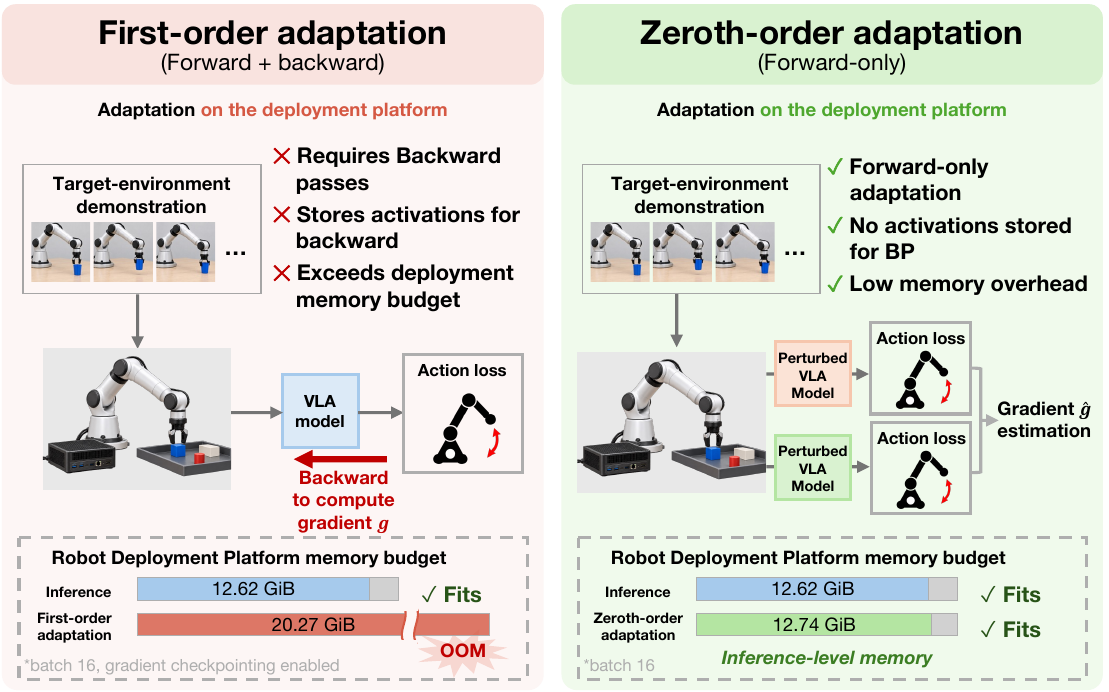}
    \caption{Adaptation on the robot deployment platform}
    \label{fig:motivation-concept}
  \end{subfigure}
  \hfill
  \begin{minipage}[b]{0.3\linewidth}
    \centering
    \begin{subfigure}[b]{\linewidth}
      \centering
      \includegraphics[width=\linewidth]{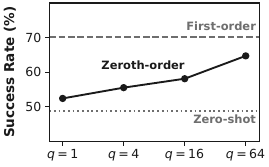}
      \caption{Adaptation quality}
      \label{fig:motivation-quality}
    \end{subfigure}\\[0.6em]
    \begin{subfigure}[b]{\linewidth}
      \centering
      \includegraphics[width=\linewidth]{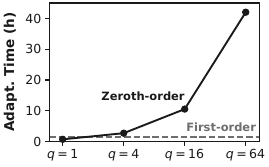}
      \caption{Adaptation time}
      \label{fig:motivation-time}
    \end{subfigure}
  \end{minipage}
  \caption{\textbf{Zeroth-order adaptation enables on-robot adaptation within the inference memory budget.} Unlike first-order adaptation, it requires only forward evaluations and avoids storing activations (a). Increasing the query budget \(q\) improves adaptation quality toward first-order performance, at the cost of increased adaptation time (b,c).
  }
  \label{fig:motivation}
\end{figure}

This adaptation, however, must run on the deployment platform itself, under a compute and memory budget provisioned for inference rather than training.
Conventional first-order adaptation fits this setting poorly (Figure~\ref{fig:motivation-concept}, left): backpropagation requires retaining intermediate activations that can otherwise be discarded during inference, substantially increasing memory requirements~\citep{3495724.3496671, Yang_2023_CVPR}.
Even with gradient checkpointing enabled, this overhead grows with batch size: for $\pi_{0.5}$, first-order adaptation uses 20.27 GiB of peak memory at batch size 16 and 54.69 GiB at batch size 64, compared with the corresponding inference-level footprints of 12.74 GiB and 27.71 GiB for forward-only ZO. The relative gap therefore increases from $1.59\times$ to $1.97\times$.
Moreover, gradient checkpointing trades memory for additional recomputation and does not remove the need for a backward-capable execution path.
These requirements make first-order adaptation difficult to accommodate in inference-oriented deployment stacks as VLA models scale.

This motivates zeroth-order (ZO) optimization as a natural alternative (Figure~\ref{fig:motivation-concept}, right), as it enables adaptation using only forward evaluations without backpropagation.
Rather than explicitly computing gradients through backpropagation, ZO methods estimate gradients by perturbing model parameters and comparing the resulting objective values using only forward evaluations~\citep{duchi2015optimal,nesterov2017random,malladi2023finetuning}.
By estimating parameter updates solely from forward evaluations, ZO avoids backpropagation and substantially reduces the memory requirements for adaptation, making it attractive for memory-constrained devices as well as deployment environments where backward execution is unavailable or undesirable.
Despite these advantages, the potential of ZO optimization for adapting VLA policies remains largely unexplored.

In this work, we first show that ZO optimization can effectively adapt VLA policies under deployment-time distribution shifts, but at a substantial runtime cost from repeated forward evaluations (Figure~\ref{fig:motivation}).
This trade-off is governed by the query budget $q$, the number of random perturbation directions evaluated per optimization step, each as a $\pm$ pair of forward passes.
A larger $q$ improves the gradient estimate, allowing adaptation quality to approach first-order performance while peak memory remains at the inference level.
The same knob, however, also determines the computational cost: every additional direction requires two more forward passes through the full model, causing adaptation time to grow with $q$ and quickly exceed that of first-order adaptation.
Because modern VLAs contain expensive vision and language components, naively applying ZO can therefore require hours of adaptation.
Thus, the main obstacle to practical ZO-based VLA adaptation is not whether it works, but whether it can run fast enough.

\section{Related Work}

\paragraph{VLA adaptation.}
Recent work has increasingly studied data-efficient adaptation of pretrained VLA policies from limited target-domain supervision~\citep{li2025controlvla,zhang2026alignthensteer,ijcai2026p569,nguyen2026foca}.
At the extreme of this low-data regime, one-shot methods adapt a VLA policy
using only a single target-domain demonstration.
FTM and FLA enable such adaptation to novel camera viewpoints~\citep{Li_2026_CVPR}, while DART addresses visual and embodiment shifts and
considers domain-level adaptation across tasks~\citep{10.1007/978-3-032-37574-2_24}.
Our focus is complementary: rather than developing a new adaptation method,
we study how to reduce the computational cost of zeroth-order VLA adaptation
under deployment-time compute and memory constraints.

\paragraph{Zeroth-order optimization.}
Zeroth-order (ZO) optimization estimates gradients from perturbed forward
evaluations without backpropagation.
MeZO~\citep{malladi2023finetuning} enables LLM fine-tuning with approximately the memory footprint of
inference and reports performance comparable to backpropagation on several
language tasks.
Subsequent benchmarking examines how task alignment and optimizer design
affect ZO fine-tuning~\citep{zhang2024revisiting}, while SubZero explores
low-rank perturbations in random subspaces~\citep{Yu_2025_ICCV}.
For execution efficiency, MobiZO parallelizes perturbation evaluations through
Multi-Perturbed LoRA within existing inference engines~\citep{gao-etal-2025-mobizo}.
In general deep-network training, DeepZero combines sparse coordinate-wise
gradient estimation with reuse of features preceding the perturbed layer
and forward parallelization~\citep{chen2024deepzero}.

\section{Zeroth-Order VLA Adaptation}

\subsection{Problem Setup and ZO Workload}
\label{sec:problem_setup}
We study post-deployment adaptation of a pretrained VLA policy from a single
target-domain demonstration, a common one-shot VLA adaptation setting used in
recent studies~\citep{Li_2026_CVPR,zhao2025mos,10.1007/978-3-032-37574-2_24}.
Following the domain-level setting of~\citet{10.1007/978-3-032-37574-2_24}, the demonstration is collected from one task in the target domain, and the adapted policy is evaluated on both the demonstrated task and unseen tasks in the same domain.
For adaptation, we use one target-domain demonstration trajectory per scene, $\mathcal{D} = \{(o_i, \ell, a_i)\}_{i=1}^{N}$
, where $N$ is the number of observation--action pairs, $o_i$ denotes the $i$-th observation (camera images and proprioceptive state), $\ell$ is the language instruction shared across the trajectory, and $a_i$ is the corresponding action chunk.

To characterize the computation overhead during ZO adaptation, we separate the VLA policy into a \emph{vision-language prefix} $f_\phi$ with parameters $\phi$ and an \emph{action expert} $g_\psi$ with parameters $\psi$.
Let $\pi_\theta$, with $\theta=(\phi,\psi)$, denote the VLA policy.
For each sample $(o_i,\ell,a_i)\in\mathcal{D}$, the frozen prefix first
forms a complete, sample-specific conditioning state $h_i$; the action
expert then uses $h_i$ to define the policy over action chunks:
\begin{equation}
\begin{aligned}
h_i &= f_\phi(o_i,\ell) && \text{(vision-language prefix)}, \\
\pi_\theta(a_i\mid o_i,\ell) &= g_\psi(a_i\mid h_i) && \text{(action expert or head)}.
\end{aligned}
\label{eq}
\end{equation}
Here, $h_i$ denotes the conditioning state passed to the action module, and $g_\psi$ denotes the action expert that maps $h_i$ to a chunk of low-level commands.
In the $\pi_{0.5}$ architecture, $h$ comprises the prefix's attention keys and values together with the mask marking its valid positions, whereas in OpenVLA-OFT, $h_i$ is represented by the action-token hidden state.
This separation is common to recent VLA designs, including OpenVLA-OFT~\citep{KimM1-RSS-25}, $\pi_0$~\citep{BlackK-RSS-25}, $\pi_{0.5}$~\citep{black2025pi}, SmolVLA~\citep{shukor2025smolvla}, and the GR00T family~\citep{bjorck2025gr00t}.
In all of them, the prefix accounts for the majority of parameters and of the floating-point operations of a single forward evaluation of the adaptation objective.

Let $\lambda$ denote the trainable adaptation parameters, which induce adapted model weights $\theta(\lambda)$, with $\theta(0) = \theta$ recovering the base policy.
For sample $i$, let $\ell_i(\lambda;\xi)$ denote the action loss of the
adapted policy, where $\xi$ collects any stochastic variables required by
the action objective.
For $\pi_{0.5}$, $\xi$ consists of an action-space noise sample and a flow
time.
The adaptation objective is
\begin{equation}
  \mathcal{L}(\lambda)
  =
  \frac{1}{N}
  \sum_{i=1}^{N}
  \mathbb{E}_{\xi}
  \left[
    \ell_i(\lambda;\xi)
  \right].
  \label{eq:objective}
\end{equation}

Zeroth-order optimization replaces the gradient of Eq.~(\ref{eq:objective}) with a randomized finite-difference estimate~\citep{duchi2015optimal, nesterov2017random}.
Let $\mathcal{L}_\mathcal{B}(\lambda; \xi)$ denote the same objective averaged over a minibatch $\mathcal{B}$ of $B$ samples drawn from $\mathcal{D}$, and evaluated at a fixed draw $\xi = (\xi_i)_{i \in \mathcal{B}}$.
With $q$ perturbation directions $z_j \sim \mathcal{N}(0, I_{d_\lambda})$,  where $d_\lambda$ is the dimension of $\lambda$ and $I_{d_\lambda}$ is the corresponding identity matrix, perturbation scale $\epsilon$, and a random draw $\xi_j$ for the
minibatch associated with each direction $j$, % ξ를 per-step보다 per-direction으로 사용해야 gradient estimation 분산이 적음 (이 내용 논문에 작성해야하나?)
\begin{equation}
  \hat{\nabla}\mathcal{L}(\lambda) \;=\; \frac{1}{q} \sum_{j=1}^{q} \frac{\mathcal{L}_\mathcal{B}(\lambda + \epsilon z_j; \, \xi_j) - \mathcal{L}_\mathcal{B}(\lambda - \epsilon z_j; \, \xi_j)}{2\epsilon} \, z_j ,
  \label{eq:rge}
\end{equation}
where the two evaluations of a $\pm$ pair share the draw $\xi_j$, and $\mathcal{B}$ is held fixed across all $q$ directions of an update.
An update therefore consumes $2q$ objective evaluations (equivalently, $2qB$ sample-level action evaluations) and never constructs a backward graph.
Two properties of this workload determine what an implementation can reuse: \emph{which} parameters adaptation has to place under the perturbation (Section~\ref{sec:placement}), and \emph{what} each of the $2qB$ evaluations is forced to recompute as a result (Section~\ref{sec:recomputes}).

\subsection{Action-Side Adaptation with a Frozen Prefix}
\label{sec:placement}

\begin{wrapfigure}{r}{0.40\textwidth}
    \vspace{-2em}
    \centering

    \begin{subfigure}[t]{0.95\linewidth}
        \centering
        \includegraphics[width=\linewidth]{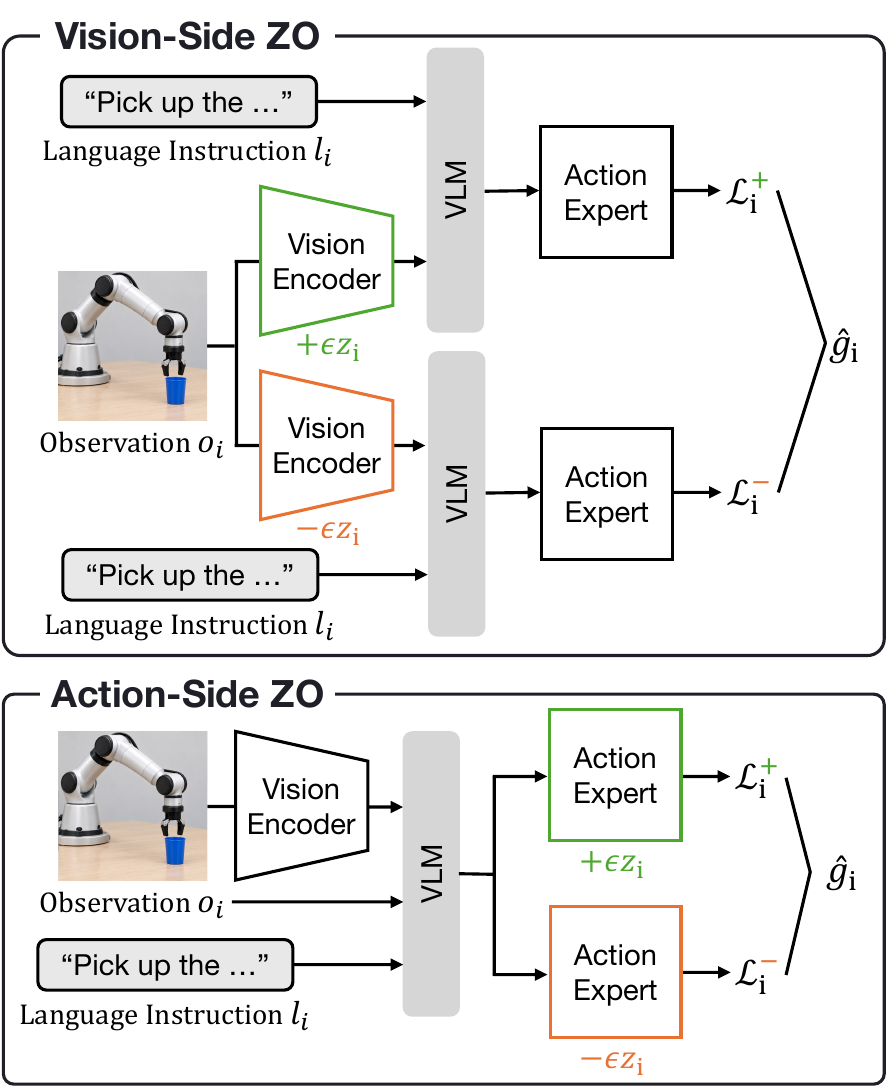}
        \caption{ZO adaptation placement.}
        \label{fig:placement}
    \end{subfigure}

    \vspace{0.5em}

    \begin{subfigure}[t]{0.95\linewidth}
        \centering
        \includegraphics[width=\linewidth]{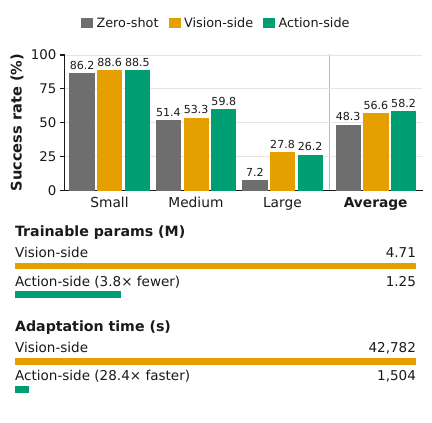}
        \caption{Viewpoint-shift experiment results.}
        \label{fig:placement_sr}
    \end{subfigure}

    \caption{
        \textbf{Action-side ZO improves success with a frozen prefix.}
    }

    \vspace{-2em}
\end{wrapfigure}

Prior VLA adaptation methods often adapt vision-side parameters, including vision encoders, visual representations, and visual projection layers~\citep{kim2024openvla,grover-2026-enhancin-iros,abouzeid2025geoaware,Li_2026_CVPR}. Among these, Feature Linear Adaptation (FLA)~\citep{Li_2026_CVPR} adapts the vision encoder with low-rank updates while keeping the downstream language and action components frozen.
We instead ask whether ZO adaptation can be confined to the action expert while keeping the expensive vision-language prefix frozen.

Figure~\ref{fig:placement} compares two LoRA placements for $\pi_{0.5}$.
The \emph{vision-side} placement targets the feed-forward projections of every block
of the vision encoder inside $f_\phi$, following the placement used by~\citet{Li_2026_CVPR}.
The \emph{action-side} placement targets the query and value projections of every
self-attention block of the action expert $g_\psi$.
At the same rank $r=16$, the action-side adapter contains $1.25$M trainable parameters,
compared with $4.71$M for the vision-side adapter.
Both configurations keep the pretrained model weights frozen and optimize only their
LoRA parameters with ZO using the objective in Eq.~(\ref{eq:objective}).

Under this matched setup, Figure~\ref{fig:placement_sr} shows that action-side ZO increases average success across the three camera-viewpoint shifts from $48.3\%$ to $58.2\%$, while matching vision-side ZO on average ($56.6\%$).
Given that it uses approximately $3.8\times$ fewer trainable parameters, this result motivates confining adaptation to the action expert while keeping the vision-language prefix frozen.

This makes action-side adaptation a more efficient choice in our method, achieving comparable performance with substantially shorter adaptation time, in addition to requiring fewer trainable parameters.
This efficiency stems from the prefix output being independent of the perturbed parameters.
For a fixed demonstration sample, its conditioning state is therefore invariant across ZO probes and optimizer steps, exposing the redundancy analyzed next.

\subsection{Repeated Prefix Computation}
\label{sec:recomputes}

Thus, each $h_i$ needs to be computed at most once and can be reused whenever the same sample is revisited, reducing the number of prefix executions from $2qBT$ to at most $N$.
In contrast, the $2qBT$ action-expert evaluations remain necessary because the action-expert parameters change across ZO probes.

Under action-side adaptation, the prefix parameters $\phi$ remain fixed, while only the action-expert parameters depend on the adaptation parameters $\lambda$, i.e., $\theta(\lambda) = (\phi, \psi(\lambda))$. Consequently, for each demonstration sample $i$, the conditioning state $h_i = f_\phi(o_i,\ell)$ is independent of $\lambda$.
A naive implementation recomputes the vision-language prefix for every one of these evaluations.
However, the demonstration contains only $N$ distinct observations $\{o_i\}_{i=1}^{N}$, and each observation has a fixed conditioning state $h_i$ because $\phi$ is frozen.
The required $2qBT$ action-expert evaluations cannot be removed, but the number of prefix executions can in principle be reduced from $2qBT$ to at most $N$ if each distinct conditioning state is computed once, cached, and reused thereafter.

The remaining systems challenge is to store and supply these states efficiently to the action expert, which Section~\ref{sec:method} addresses through within-step reuse, across-step reuse, and schedule-aware prefetch.

\section{\system{}: Reusing Frozen Conditioning Across ZO Probes}
\label{sec:method}

\begin{figure}[H]
  \centering
  \includegraphics[width=\linewidth]{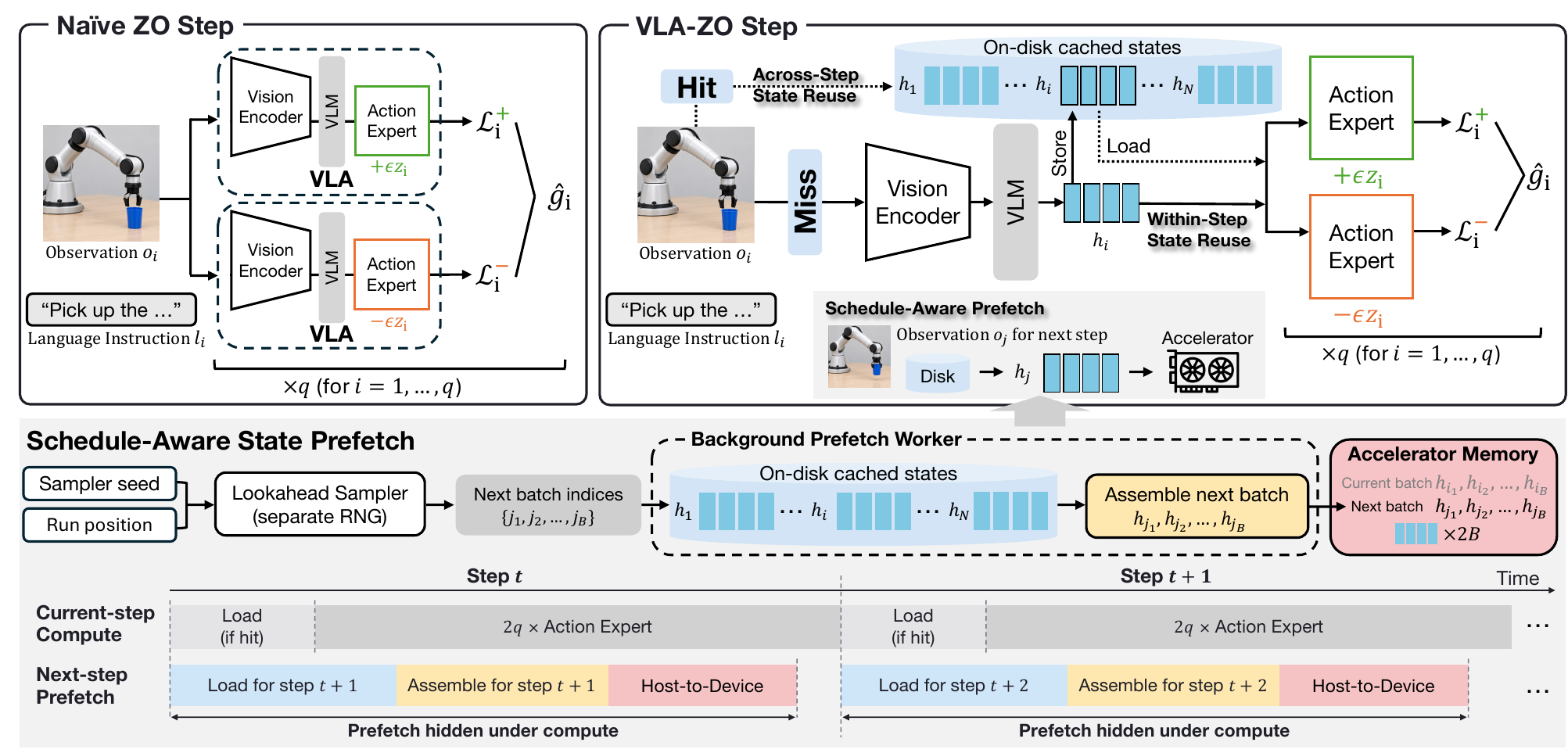}
  \caption{Overview of \system{}.
  }
  \label{fig:method-overview}
\end{figure}

\system{} aims to accelerate ZO adaptation without changing its objective, perturbation directions, or trainable parameter space.
The key observation is that once adaptation is confined to the action expert, the expensive vision-language prefix becomes independent of the ZO perturbation.
In a naive ZO step, each of the $2q$ perturbed evaluations recomputes the same
frozen-prefix state for a sample, resulting in $2qBT$ prefix executions over
$T$ updates despite only $N$ distinct conditioning states.
\system{} addresses this computational bottleneck in ZO adaptation through three optimizations, illustrated in Figure~\ref{fig:method-overview}.
First, within an update step, it computes the prefix once per sample and reuses the resulting state across all \(2q\) probes (Section~\ref{sec:prefix-reuse}).
Second, the same prefix states are retained across update steps, so that each distinct observation requires only one prefix evaluation over the full adaptation run (Section~\ref{sec:across-step-reuse}).
Finally, \system{} overlaps transfer of the next cached batch with the current probe evaluations using schedule-aware prefetching (Section~\ref{sec:cache-prefetch}).

\paragraph{Cost model.}
Let $c_{\mathrm{pre}}$ and $c_{\mathrm{suf}}$ denote the per-sample costs of the frozen prefix and action suffix, respectively, and let $c_{\mathrm{ld}}$ denote the per-sample cost of loading a cached conditioning state.
For comparison, we separate the recurring per-update cost from the one-time cost of materializing the $N$ distinct conditioning states, which is $N c_{\mathrm{pre}}$.
The resulting per-update costs are
\begin{subequations}
\label{eq:costs}
\begin{align}
  C_{\mathrm{naive}}
  &= B\bigl(2q\,c_{\mathrm{suf}} + 2q\,c_{\mathrm{pre}}\bigr),
  \label{eq:cost-naive}\\
  C_{\mathrm{within}}
  &= B\bigl(2q\,c_{\mathrm{suf}} + c_{\mathrm{pre}}\bigr),
  \label{eq:cost-within}\\
  C_{\mathrm{across}}
  &= B\bigl(2q\,c_{\mathrm{suf}} + c_{\mathrm{ld}}\bigr),
  \label{eq:cost-across}\\
  \text{and}\quad C_{\mathrm{prefetch}}
  &= B\bigl(2q\,c_{\mathrm{suf}}
  + \max(0,\; c_{\mathrm{ld}} - 2q\,c_{\mathrm{suf}})\bigr).
  \label{eq:cost-prefetch}
\end{align}
\end{subequations}
Here, $C_{\mathrm{naive}}$ denotes the per-update cost of naive execution, while $C_{\mathrm{within}}$, $C_{\mathrm{across}}$, and $C_{\mathrm{prefetch}}$ denote the costs after progressively adding within-step reuse, across-step reuse, and schedule-aware prefetch, respectively. The common term $2qB c_{\mathrm{suf}}$ is the action-expert work required by the ZO estimator in Eq.~(\ref{eq:rge}).
The three mechanisms below progressively reduce the remaining prefix-related overhead: first by eliminating repeated prefix execution, then by reusing cached prefix states across update steps, and finally by hiding cache-transfer latency behind action-expert computation.

\subsection{Within-Step State Reuse}
\label{sec:prefix-reuse}

Within a ZO update, all $2q$ probes for a given sample share the same input $(o_i,\ell)$.
Because adaptation is confined to the action expert, perturbing the adaptation parameters changes only the action expert, while the frozen prefix produces the same conditioning state $h_i$ for every probe.
\system{} computes $h_i=f_\phi(o_i,\ell)$ once per sample and reuses it across all $2q$ perturbed action-expert evaluations.
For a minibatch of size $B$, this reduces the number of prefix evaluations per update from $2qB$ to $B$, while leaving the required $2qB$ action-expert evaluations unchanged.
Accordingly, the per-update cost decreases from Eq.~(\ref{eq:cost-naive}) to Eq.~(\ref{eq:cost-within}).

The mathematical equivalence of reuse and recomputation rests on three conditions, all of which are met by our cached action-side implementations.
First, all perturbed parameters must lie after the reuse boundary: we adapt the action expert in $\pi_{0.5}$ and the action head in OpenVLA-OFT.
Second, $h_i$ must contain the complete information passed from the prefix to the action expert: $\pi_{0.5}$ caches the prefix attention
  state and padding mask, while OpenVLA-OFT caches the action-token hidden features consumed by its action head.
Third, the frozen prefix must be deterministic for a fixed $(o_i,\ell)$ under the execution configuration used for caching.
Thus, under exact arithmetic, reuse leaves the objective and finite-difference estimator unchanged.

\paragraph{Effect of the query budget.}
Without reuse, each additional perturbation direction adds two full-policy evaluations per sample, with cost $2(c_{\mathrm{pre}}+c_{\mathrm{suf}})$.
With within-step reuse, it adds only two action-expert evaluations, with cost $2c_{\mathrm{suf}}$.
Relative to naive execution, the per-update speedup of within-step reuse is
\begin{equation}
  \frac{C_{\mathrm{naive}}}{C_{\mathrm{within}}}
  = \frac{2q(c_{\mathrm{pre}}+c_{\mathrm{suf}})} {c_{\mathrm{pre}}+2q c_{\mathrm{suf}}}.
  \label{eq:within-speedup}
\end{equation}
As $q$ increases, the prefix cost is amortized over more probes and the speedup approaches $(c_{\mathrm{pre}}+c_{\mathrm{suf}})/c_{\mathrm{suf}}$, the cost ratio between a full forward evaluation and an action-expert-only evaluation.
This is useful in the large-$q$ regime where ZO obtains a better gradient estimate but naive execution becomes increasingly expensive.

\subsection{Across-Step State Reuse}
\label{sec:across-step-reuse}
Within-step reuse still recomputes the prefix whenever a sample appears in a new optimizer step.
In one-shot adaptation, however, every minibatch is drawn from the same demonstration of $N$ observations, and the prefix remains frozen for the entire run.
Therefore, once $h_i$ has been computed for a sample, the same state can be reused every time that sample is revisited.
\system{} stores each conditioning state after its first computation and looks it up on subsequent visits.
The cache is populated lazily, so a miss runs the prefix once and stores the result, while a hit skips prefix execution entirely.
Over $T$ updates, within-step reuse performs $BT$ prefix evaluations, whereas ideal across-step reuse requires at most one prefix evaluation for each of the $N$ demonstration observations.
All adaptation times reported in Section~\ref{sec:experiments} include the cost of populating the cache on first visits.

Once all $N$ demonstration observations have been encountered, the cumulative execution costs over $T$ updates are
\begin{equation}
\begin{aligned}
C_{\mathrm{within}}(T)
&=
BT\left(c_{\mathrm{pre}} + 2q c_{\mathrm{suf}}\right),\\
C_{\mathrm{across}}(T)
&=
N c_{\mathrm{pre}}
+
(BT-N)c_{\mathrm{ld}}
+
2qBTc_{\mathrm{suf}}.
\end{aligned}
\label{eq:across-step-cost}
\end{equation}

\paragraph{Effect of repeated visits.}
Across-step reuse replaces repeated prefix executions with cached-state loads.
Relative to within-step reuse, the cumulative cost reduction over $T$ update steps is
\begin{equation}
C_{\mathrm{within}}(T)-C_{\mathrm{across}}(T)
=
(BT-N)\left(c_{\mathrm{pre}}-c_{\mathrm{ld}}\right).
\label{eq:across-saving}
\end{equation}
Thus, the benefit grows with the number of repeated sample visits, $BT-N$, while the one-time cost of materializing the $N$ distinct states is amortized over more updates.
As discussed in Section~\ref{sec:problem_setup}, the vision-language prefix accounts for the majority of the forward-pass FLOPs in most VLA architectures.
Consequently, loading a cached conditioning state is substantially cheaper than recomputing the prefix, with $c_{\mathrm{ld}} \ll c_{\mathrm{pre}}$.

\subsection{Schedule-Aware State Prefetch}
\label{sec:cache-prefetch}

Across-step reuse replaces repeated prefix execution with cached-state loading,
but keeping all $N$ conditioning states resident on the accelerator would make
device memory grow with the demonstration size.
\system{} therefore stores the persistent cache on disk and stages only the
states required by each minibatch.
This bounds accelerator-side cache storage, but places disk loading, batch
assembly, and host-to-device transfer on the critical path if performed
synchronously.
\system{} hides this latency by overlapping preparation of the next cached
batch with the $2q$ probe evaluations of the current update, as illustrated in
Figure~\ref{fig:method-overview}.

\paragraph{Schedule-aware lookahead.}
Let $\mathcal{B}_t$ denote the minibatch used at update $t$ and let $\kappa_t$
denote the corresponding sampler position.
For a fixed sampler seed $s$, the minibatch schedule is determined by the
sampling rule, $\mathcal{B}_t = \mathcal{S}(s,\kappa_t)$.
Thus, while update $t$ is running, \system{} can reconstruct
$\mathcal{B}_{t+1}$ from $(s,\kappa_{t+1})$ using a separate local generator.
This lookahead follows the same permutation and batch-formation rules as the
live sampler, but does not advance its state or consume randomness used for
the perturbations or objective.

For cached samples in $\mathcal{B}_{t+1}$, a background worker reads the
conditioning states, assembles the next batch, and transfers it to the
accelerator while the current batch evaluates its $2q$ action-expert probes.
The worker never invokes the policy or materializes cache misses; an
incompletely cached future batch remains on the ordinary demand path when that
update begins.
We prefetch only one update ahead, so device staging contains at most the
current and next batches of conditioning states.
Implementation details are given in Appendix~\ref{app:exactness}.

\paragraph{Effect of latency hiding.}
For a fully cached minibatch, loading its conditioning states costs
$B c_{\mathrm{ld}}$, while evaluating the $2q$ action-expert probes costs
$2qB c_{\mathrm{suf}}$.
Without prefetching, these costs are serialized, yielding
$C_{\mathrm{across}} = B(c_{\mathrm{ld}} + 2q c_{\mathrm{suf}})$.
With one-update lookahead, loading for the next update overlaps with probe
computation for the current update.
After pipeline warm-up, the recurring per-update cost is therefore determined
by the slower of the two:
\begin{align}
    C_{\mathrm{prefetch}}
    &= \max(Bc_{\mathrm{ld}},\, 2qB c_{\mathrm{suf}}) \nonumber\\
    % &= B\max(c_{\mathrm{ld}},\, 2q c_{\mathrm{suf}}) \nonumber\\
    &= B\left(
        2q c_{\mathrm{suf}}
        + \max(0, c_{\mathrm{ld}} - 2q c_{\mathrm{suf}})
       \right),
\end{align}
recovering Eq.~(\ref{eq:cost-prefetch}).
Thus, the loading latency that remains exposed on the critical path is
$B\max(0, c_{\mathrm{ld}} - 2q c_{\mathrm{suf}})$.
Loading is fully hidden whenever $c_{\mathrm{ld}} \le 2q\,c_{\mathrm{suf}}$,
or, equivalently, when $q \ge \left\lceil c_{\mathrm{ld}}/(2c_{\mathrm{suf}}) \right\rceil$.
Thus, increasing the query budget also lengthens the suffix-computation window
available to hide cache I/O.
Once the condition above holds, the recurring update cost reduces to
$2qB c_{\mathrm{suf}}$, and the speedup relative to naive execution reaches
$(c_{\mathrm{pre}}+c_{\mathrm{suf}})/c_{\mathrm{suf}}$.

\section{Experiments}
\label{sec:experiments}

\subsection{Setup}

\paragraph{Benchmark and Evaluation}
We evaluate $\pi_{0.5}$~\citep{black2025pi} and
OpenVLA-OFT~\citep{KimM1-RSS-25} on all 40 tasks in the LIBERO Spatial,
Object, Goal, and LIBERO-10 suites~\citep{liu2023libero}.
Following DART's scene-wise one-shot protocol~\citep{10.1007/978-3-032-37574-2_24},
adaptation uses five expert demonstrations, one per scene, rendered under
the target visual condition. We test three camera shifts
(\texttt{Small}, \texttt{Medium}, and \texttt{Large}) and compound shifts
combining \texttt{Medium} with image blur or altered lighting.
Each condition is evaluated over 40 initial states per task (1{,}600 episodes), using indices 10--49 and including tasks without adaptation demonstrations. Initial-state indices 0--9 are excluded because they were used for learning-rate selection.

\paragraph{Baselines and Ablations}
We compare \system{} with the unadapted policy,
FLA~\citep{Li_2026_CVPR}, and
DART~\citep{10.1007/978-3-032-37574-2_24}.
Our FLA and DART baselines, denoted by $\dagger$, use zeroth-order
gradient estimates computed using only forward passes
and update the parameters with SGD.
At each query budget, the execution ablation fixes the adaptation settings
and progressively enables within-step reuse, across-step reuse, and
schedule-aware prefetch. Appendix~\ref{app:experiments} provides the data
protocol, perturbations, model-specific rollout settings, hyperparameters,
and timing procedure.

\begin{table}[t]
\caption{\textbf{Camera-viewpoint adaptation of $\pi_{0.5}$.}
At fixed \(q\), successive action-side rows add execution optimizations under identical adaptation settings. Times include cold-cache population; speedups are relative to baseline ZO at the same \(q\). † denotes our ZO-SGD baseline implementations.}
\label{tab:main}
\centering\fontsize{8}{9.6}\selectfont
\setlength{\tabcolsep}{3pt}
\begin{tabular}{lcccccccc}
\toprule
& & \multicolumn{4}{c}{Novel viewpoint (success rate, \%)} & \multicolumn{3}{c}{Adaptation} \\
\cmidrule(lr){3-6}\cmidrule(lr){7-9}
Method & $q$ & \texttt{Small} & \texttt{Medium} & \texttt{Large} &
Average & Params (M) & Time (s) & Speedup \\
\midrule
Zero-shot & -- & 86.19 & 51.44 & 7.19 & 48.27 & -- & -- & -- \\
FLA$^\dagger$~\citep{Li_2026_CVPR}
& 16 & 88.56 & 53.31 & 27.81 & 56.56 & 4.71 & 42,782 & -- \\
DART$^\dagger$~\citep{10.1007/978-3-032-37574-2_24}
& 16 & 89.50 & 58.88 & 14.25 & 54.21 & 3,353.43 & 43,794 & -- \\
\midrule
Baseline ZO
& 16 & 86.44 & 60.69 & 25.75 & 57.63 &1.25
& 38,487 & 1.00$\times$ \\
\quad + Within-step prefix reuse
& 16 & 88.00 & 60.31 & 25.00 & 57.77 &1.25
& 2,474 & 15.56$\times$ \\
\quad + Across-step prefix reuse
& 16 & 88.06 & 60.88 & 25.56 & 58.17 &1.25
& 2,264 & 17.00$\times$ \\
\hlc\quad + Schedule-aware prefetch (\system{})
& \hlc 16 & \hlc 88.50 & \hlc 59.81 & \hlc 26.19 & \hlc 58.17
& \hlc 1.25 & \hlc 1,504 & \hlc 25.59$\times$ \\
\midrule
Baseline ZO
& 64 & 87.31 & 67.00 & 37.00 & 63.77 &1.25
& 153,220 & 1.00$\times$ \\
\quad + Within-step prefix reuse
& 64 & 87.56 & 66.38 & 36.63 & 63.52 &1.25
& 5,734 & 26.72$\times$ \\
\quad + Across-step prefix reuse
& 64 & 87.19 & 65.75 & 35.56 & 62.83 &1.25
& 5,465 & 28.04$\times$ \\
\hlc\quad + Schedule-aware prefetch (\system{})
& \hlc 64 & \hlc 87.56 & \hlc 66.38 & \hlc 36.81 & \hlc 63.58
& \hlc 1.25 & \hlc 4,709 & \hlc 32.54$\times$ \\
\bottomrule
\end{tabular}
\end{table}

\begin{table}[h]
\caption{\textbf{Compound visual shifts with $\pi_{0.5}$, by LIBERO suite.} Avg. is the mean across LIBERO suites. $\dagger$ denotes our ZO-SGD baseline implementations.}
\label{tab:combined}
\centering
\fontsize{8}{9.6}\selectfont
\setlength{\tabcolsep}{3pt}
\begin{tabular}{l c c c c c @{\hspace{6pt}} c c c c c @{\hspace{6pt}} c c c}
\toprule
& & \multicolumn{5}{c}{\texttt{View+Noise}} & \multicolumn{5}{c}{\texttt{View+Light}} & \multicolumn{2}{c}{Adaptation} \\
\cmidrule(lr){3-7}\cmidrule(lr){8-12}\cmidrule(lr){13-14}
Method & $q$ & Spa. & Obj. & Goal & 10 & \textbf{Avg.} & Spa. & Obj. & Goal & 10 & \textbf{Avg.} & Params (M) & Time (s) \\
\midrule
Zero-shot & -- & 50.5 & 84.0 & 42.0 & 20.3 & 49.2 & 39.8 & 86.8 & 53.8 & 15.8 & 49.0 & -- & -- \\
FLA$^\dagger$~\citep{Li_2026_CVPR} & 16 & 65.3 & 70.8 & 47.0 & 29.0 & 53.0 & 43.5 & 73.5 & 57.5 & 35.3 & 52.4 & 4.71 & 43,081 \\
DART$^\dagger$~\citep{10.1007/978-3-032-37574-2_24} & 16 & 57.5 & 90.8 & 45.5 & 27.8 & 55.4 & 52.3 & 92.3 & 50.8 & 31.8 & 56.8 & 3,353.43 & 43,965 \\
\hlc\system{}
& \hlc 16
& \hlc 46.0 & \hlc 86.8 & \hlc 43.5 & \hlc 28.0 & \hlc 51.1
& \hlc 60.0 & \hlc 90.3 & \hlc 56.3 & \hlc 27.0 & \hlc 58.4
& \hlc 1.25 & \hlc 1,500 \\

\hlc\system{}
& \hlc 64
& \hlc 49.0 & \hlc 91.5 & \hlc 49.3 & \hlc 33.3 & \hlc 55.8
& \hlc 68.5 & \hlc 89.0 & \hlc 59.0 & \hlc 26.8 & \hlc 60.8
& \hlc 1.25 & \hlc 4,726 \\
\bottomrule
\end{tabular}
\end{table}

\subsection{Camera-Viewpoint Adaptation}
Table~\ref{tab:main} evaluates two aspects of \system{}: whether adapting the
action expert improves robustness to a changed viewpoint, and how much time
is saved by executing a fixed ZO workload with progressively more reuse.
The zero-shot, FLA, and DART rows provide adaptation-quality reference points;
the four action-side rows at each $q$ provide a cumulative ablation of the execution mechanisms.

\paragraph{Adaptation quality and query budget.}
The action-side adapter also improves average success with fewer trainable
parameters than either ZO baseline.
At $q=16$, \system{} reaches $58.17\%$ average success, compared with
$48.27\%$ for the unadapted policy. At $q=64$, its average rises to
$63.58\%$. The larger budget helps most under the \texttt{Medium} and
\texttt{Large} shifts, where success increases from $59.81\%$ to $66.38\%$
and from $26.19\%$ to $36.81\%$, respectively. The \texttt{Small} condition
already has high zero-shot success and does not improve consistently with
$q$. Thus, additional perturbation directions are most useful in the harder
viewpoint conditions in this experiment.

\paragraph{Execution time and the contribution of reuse.}
At $q=16$, the full system reduces adaptation time from 38{,}487 to
1{,}504 seconds ($25.59\times$). At $q=64$, it reduces time from 153{,}220
to 4{,}709 seconds ($32.54\times$). Within-step reuse accounts for the
largest reduction, bringing the two workloads to 2{,}474 and 5{,}734 seconds.
Across-step reuse then lowers these times to 2{,}264 and 5{,}465 seconds,
and prefetching supplies the final reduction. This ordering is consistent
with the cost model in Eq.~(\ref{eq:costs}): first remove prefix execution
from individual probes, then reuse conditioning across updates, and finally
overlap its loading with action-expert computation.

\paragraph{Quality across execution variants.}
At fixed $q$, reuse changes when conditioning is computed and transferred,
while preserving the optimization problem. The observed average-success
ranges are $57.63$--$58.17\%$ at $q=16$ and $62.83$--$63.77\%$ at $q=64$.
These variations should not be interpreted as quality improvements or degradations from
caching. Split prefix/suffix execution can change floating-point kernels,
and small loss differences can accumulate through finite differences and
successive updates (Section~\ref{sec:reuse-preservation}). The table supports
similar observed adaptation quality.

\subsection{Compound Visual Shifts}
Table~\ref{tab:combined} examines whether the
benefits extend beyond camera geometry. Each compound condition retains the
\texttt{Medium} viewpoint and adds blur or altered lighting.
Under \texttt{View+Noise}, \system{} reaches $51.1\%$ average success at
$q=16$ and $55.8\%$ at $q=64$, compared with $49.2\%$ without adaptation.
Under \texttt{View+Light}, the corresponding averages are $58.4\%$ and
$60.8\%$, compared with $49.0\%$ without adaptation. The gains are larger
for the lighting condition. At $q=64$, \system{} is close to the DART
baseline on \texttt{View+Noise} and exceeds it on \texttt{View+Light}.

\subsection{Generalization to OpenVLA-OFT}
OpenVLA-OFT places a separate L1 action-regression head after the frozen
vision-language model. It evaluates reuse at a different architectural
boundary from the flow-matching action expert of $\pi_{0.5}$.

\begin{table}[t]
\caption{\textbf{Camera-viewpoint adaptation of OpenVLA-OFT.}
Success rates (\%) are reported by viewpoint severity and averaged across
severities. Action-side adaptation targets the separate L1 regression head.
The reported \system{} timing includes cold feature-cache population.
$\dagger$ denotes our ZO-SGD baseline implementations.}
\label{tab:main_openvla}
\centering\fontsize{8}{9.6}\selectfont
\setlength{\tabcolsep}{3pt}
\begin{tabular}{lccccccc}
\toprule
& & \multicolumn{4}{c}{Novel viewpoint (success rate, \%)} &
\multicolumn{2}{c}{Adaptation} \\
\cmidrule(lr){3-6}\cmidrule(lr){7-8}
Method & $q$ & \texttt{Small} & \texttt{Medium} & \texttt{Large} &
Average & Params (M) & Time (s) \\
\midrule
Zero-shot
& -- & 80.19 & 50.88 & 17.38 & 49.48
& -- & -- \\

FLA$^\dagger$~\citep{Li_2026_CVPR}
& 4 & 81.06 & 52.63 & 19.31 & 51.00
& 4.71 & 18,283 \\

DART$^\dagger$~\citep{10.1007/978-3-032-37574-2_24}
& 4 & 81.13 & 53.63 & 20.50 & 51.75
& 7,709.17 & 28,482 \\

% \midrule
\hlc\system{}
& \hlc 4
& \hlc 79.06
& \hlc 57.06
& \hlc 21.88
& \hlc 52.67
& \hlc 0.79
& \hlc 109 \\

\hlc\system{}
& \hlc 16
& \hlc 79.63
& \hlc 55.69
& \hlc 21.25
& \hlc 52.19
& \hlc 0.79
& \hlc 170\\

\bottomrule
\end{tabular}
\end{table}

Table~\ref{tab:main_openvla} shows that action-head adaptation at $q=4$
raises average viewpoint-shift success from $49.48\%$ to $52.67\%$.
The improvement is concentrated in the \texttt{Medium} and \texttt{Large}
conditions, while \texttt{Small} changes little.
Table~\ref{tab:combined_openvla_suite} further shows that the benefit extends to compound visual shifts. Under \texttt{View+Noise}, average success increases from $23.9\%$ without adaptation to $29.2\%$ at $q=16$ and $33.5\%$ at $q=64$. Under \texttt{View+Light}, it increases from $40.9\%$ to $43.8\%$ and $46.3\%$, respectively. Together with the viewpoint results, these results show that the action-head adaptation strategy transfers to OpenVLA-OFT.

\begin{table}[t]
\caption{\textbf{Compound visual shifts with OpenVLA-OFT, by LIBERO suite.} Avg. is the mean across LIBERO suites. $\dagger$ denotes our ZO-SGD baseline implementations.}
\label{tab:combined_openvla_suite}
\centering
\fontsize{8}{9.6}\selectfont
\setlength{\tabcolsep}{3pt}
\begin{tabular}{l c c c c c @{\hspace{6pt}} c c c c c @{\hspace{6pt}} c c c}
\toprule
& & \multicolumn{5}{c}{\texttt{View+Noise}} & \multicolumn{5}{c}{\texttt{View+Light}} & \multicolumn{2}{c}{Adaptation} \\
\cmidrule(lr){3-7}\cmidrule(lr){8-12}\cmidrule(lr){13-14}
Method & $q$ & Spa. & Obj. & Goal & 10 & \textbf{Avg.} & Spa. & Obj. & Goal & 10 & \textbf{Avg.} & Params (M) & Time (s) \\
\midrule
Zero-shot & -- & 17.0 & 41.0 & 17.3 & 20.3 & 23.9 & 26.3 & 86.8 & 26.0 & 24.5 & 40.9 & -- & -- \\
FLA$^\dagger$~\citep{Li_2026_CVPR} & 4 & 21.8 & 47.3 & 14.0 & 21.8 & 26.2 & 26.5 & 87.3 & 25.8 & 28.3 & 41.9 & 4.71 & 18,283 \\
DART$^\dagger$~\citep{10.1007/978-3-032-37574-2_24} & 4 & 29.8 & 43.3 & 18.0 & 24.8 & 28.9 & 30.8 & 86.0 & 24.8 & 26.8 & 42.1 & 7,709.17 & 28,265 \\

% \hlc\system{}
% & \hlc 4
% & \hlc   & \hlc   & \hlc   & \hlc   & \hlc  
% & \hlc   & \hlc   & \hlc   & \hlc   & \hlc  
% & \hlc 0.79 & \hlc  107 \\

\hlc\system{}
& \hlc 16
& \hlc 29.3 & \hlc 45.8 & \hlc 17.5 & \hlc 24.3 & \hlc 29.2
& \hlc 31.5 & \hlc 88.0 & \hlc 28.8 & \hlc 26.8 & \hlc 43.8
& \hlc 0.79 & \hlc 170 \\

\hlc\system{}
& \hlc 64
& \hlc 31.8 & \hlc 65.5 & \hlc 13.3 & \hlc 23.5 & \hlc 33.5
& \hlc 34.8 & \hlc 89.3 & \hlc 32.8 & \hlc 28.3 & \hlc 46.3
& \hlc 0.79 & \hlc 403 \\

\bottomrule
\end{tabular}
\end{table}

\section{Conclusion}
We presented \system{}, a fast forward-only framework for one-shot VLA adaptation that eliminates redundant computation across zeroth-order queries. More broadly, our results highlight a unique opportunity in ZO adaptation: speed can translate directly into adaptation quality, because faster execution enables more query directions and thus better gradient estimates within a fixed time budget. This suggests fast ZO as a promising path toward practical VLA adaptation under resource-constrained deployment platforms.

\bibliography{iclr2027_conference}
\bibliographystyle{iclr2027_conference}

\appendix
\newpage
\section{Experimental Protocol and Hyperparameters}
\label{app:experiments}

\subsection{DART Adaptation Data Protocol}
\label{app:data-protocol}

\paragraph{Scene-wise supervision and transfer across tasks.}
We adopt the task and demonstration selection protocol in Appendix C.2 and
Table 13 of DART~\citep{10.1007/978-3-032-37574-2_24}.
The four evaluation suites partition 40 tasks, whereas the five scenes
define the one-shot supervision budget.
Table~\ref{tab:adaptation-tasks} lists the first task combination, used in
the experiments here. For condition-specific adaptation, its five
trajectories are pooled into one dataset.
Evaluation includes all 40 tasks,
so the policy must transfer the adaptation to tasks outside this set of
five demonstration tasks. The demonstrations retain their task
instructions and actions across camera views.

\begin{table}[h]
\centering\small
\caption{Adaptation task combination 1 from DART's Table 13.
Lengths count retained observation-action pairs after no-op removal.}
\label{tab:adaptation-tasks}
\setlength{\tabcolsep}{4pt}
\begin{tabular}{lp{0.75\linewidth}r}
\toprule
Scene & Demonstrated task & Length \\
\midrule
Living room & Put both the alphabet soup and the tomato sauce in the basket. & 336 \\
Kitchen & Turn on the stove and put the moka pot on it. & 272 \\
Study & Pick up the book and place it in the back compartment of the caddy. & 234 \\
Floor & Pick up the alphabet soup and place it in the basket. & 145 \\
Tabletop & Pick up the black bowl between the plate and the ramekin and place it on the plate. & 98 \\
\bottomrule
\end{tabular}
\end{table}

DART defines three task combinations. The retained dataset reconstructs
all three, containing 1{,}085, 1{,}044, and 1{,}025 frames per camera view,
respectively. We use combination 1
and seed 0.

\paragraph{Selecting the first demonstration.}
The source is the filtered \texttt{openvla/modified\_libero\_}\allowbreak\texttt{rlds} dataset,
which DART uses for its demonstration selection. The first demonstration
is the lowest original per-task demonstration ID still present in this
filtered release, rather than the first serialized TFRecord encountered.
This distinction matters because shard order is shuffled and filtering
removes some original demonstrations. The implementation identifies each
eligible episode by matching its float32 action sequence to the original
LIBERO demonstration after applying the same no-op removal rule. It then
selects the lowest eligible ID. For example, the living-room trajectory in
combination 1 is \texttt{demo\_1}, because \texttt{demo\_0} is absent from
the filtered source. The selected actions for combination 1 match the
released \texttt{SNUMPR/dart-libero-viewpoints} data; the public release
itself contains only this combination.

\subsection{Visual Perturbations}
\label{app:visual-shifts}

Table~\ref{tab:camera-settings} specifies the external-camera translations.
The simulator also changes camera orientation to look toward the scene
origin, so the rotation depends on the original scene-specific camera
pose. The wrist camera is not moved by this viewpoint transformation.

\begin{table}[t]
\centering\small
\caption{Camera translations relative to the original external camera,
following the pinned LIBERO-view implementation of DART's viewpoint shifts.}
\label{tab:camera-settings}
\begin{tabular}{lrrr}
\toprule
Severity & $\Delta x$ (m) & $\Delta y$ (m) & $\Delta z$ (m) \\
\midrule
\texttt{Small} & 0.0 & 0.3 & $-0.1$ \\
\texttt{Medium} & $-0.2$ & 0.7 & $-0.2$ \\
\texttt{Large} & $-1.2$ & 1.0 & $-0.2$ \\
\bottomrule
\end{tabular}
\end{table}

The condition named \texttt{Noise} applies OpenCV Gaussian blur with a
$27\times27$ kernel and automatically inferred standard deviation
(\texttt{sigmaX=0}) to the external image. The implementation does
not blur the wrist image. \texttt{Light} loads the modified LIBERO scene
assets with stronger blue illumination and changed overhead-light
positions; scene lighting affects both camera views. Compound conditions
apply these changes together with the \texttt{Medium} camera pose.
Matching perturbed adaptation bundles preserve the demonstration actions,
task selection, and scene protocol. Action-side and FLA adapters are
trained separately for each condition. DART uses a matching target stage
and the canonical source stage. Thus, these experiments measure adaptation
with target-condition demonstrations rather than merely testing a
clean-view adapter on an additional corruption.

\subsection{Models, Action Windows, and Evaluation}
\label{app:model-settings}

Table~\ref{tab:rollout-settings} summarizes the rollout settings.
The replanning interval is the number of actions executed from each
predicted chunk before querying the policy with a new observation.
Maximum control horizons exclude the initial settling steps and are held
fixed within each model's comparisons.

\begin{table}[t]
\centering\small
\caption{Model-specific LIBERO rollout settings. The relation to released
model settings and evaluation defaults is specified below.}
\label{tab:rollout-settings}
\setlength{\tabcolsep}{8pt}
\begin{tabular}{lcc}
\toprule
Setting & $\pi_{0.5}$ & OpenVLA-OFT \\
\midrule
Predicted action chunk length & 50 & 8 \\
Actions executed before replanning & 10 & 8 \\
Maximum control steps: Spatial & 220 & 220 \\
Maximum control steps: Object & 280 & 280 \\
Maximum control steps: Goal & 300 & 300 \\
Maximum control steps: LIBERO-10 & 520 & 520 \\
Initial settling steps & 10 & 10 \\
Episodes used for LR selection & 10 (0--9) & 10 (0--9) \\
Evaluation episodes per task & 40 (10--49) & 40 (10--49) \\
Rendered camera resolution & $256\times256$ & $256\times256$ \\
Policy image resolution & $224\times224$ & $224\times224$ \\
\bottomrule
\end{tabular}
\end{table}

\paragraph{$\pi_{0.5}$ model.}
The checkpoint is \texttt{lerobot/pi05\_libero\_finetuned\_v044}.
We retain the released checkpoint's 50-action chunks, ten flow-integration
steps at inference, and pretrained observation/action preprocessing and
normalization. Images are rendered at $256\times256$, matching the
checkpoint's declared camera inputs, and resized internally to
$224\times224$. The ten-action replanning interval follows
\href{https://github.com/huggingface/lerobot/blob/7e241bd630a3719a56157a497ce5d08f244784f1/docs/source/libero.mdx}{LeRobot's documented LIBERO evaluation command};
this overrides the checkpoint's stored \texttt{n\_action\_steps=50}.
The per-suite rollout limits follow the standard OpenPI/OpenVLA LIBERO evaluation convention.
Benchmark initial-state selection and ten
settling steps match
\href{https://github.com/huggingface/lerobot/blob/7e241bd630a3719a56157a497ce5d08f244784f1/src/lerobot/envs/libero.py}{LeRobot v0.6.1's LIBERO defaults}.
The adaptation objective is the native flow-matching loss. Action noise
and flow time are resampled between perturbation directions and replayed
within each positive/negative pair. All directions in an update use the
same effective minibatch.

\paragraph{OpenVLA-OFT model.}
The checkpoint is
\texttt{moojink/}\allowbreak\texttt{openvla-7b-oft-finetuned-}\allowbreak\texttt{libero-spatial-object-goal-10}.
We follow the
\href{https://github.com/moojink/openvla-oft/blob/e4287e94541f459edc4feabc4e181f537cd569a8/experiments/robot/libero/run_libero_eval.py}{official LIBERO evaluation defaults}
for the continuous L1-regression head, two camera inputs with
proprioception, eight-action replanning, per-suite rollout limits, ten
settling steps, and 50 episodes per task. We also retain its
$256\times256$ rendering and official image preprocessing to
$224\times224$ with center cropping. The eight-action training chunks
and checkpoint-provided, suite-specific \texttt{BOUNDS\_Q99}
normalization follow the
\href{https://github.com/moojink/openvla-oft/blob/e4287e94541f459edc4feabc4e181f537cd569a8/prismatic/vla/constants.py}{released LIBERO model configuration}.
This action-head loss does not sample a flow time or action noise.

\paragraph{Evaluation and aggregation.}
For both models, we use 50 benchmark initial states per task, reserving indices 0--9 for learning-rate selection and evaluating the final models on indices 10--49. This yields 40 evaluation episodes per task, 400 per suite, and 1{,}600 episodes in total for each condition.
For $\pi_{0.5}$, this increases the ten episodes per task in LeRobot's
documented evaluation protocol to 50; for OpenVLA-OFT, it matches the
official default. We set the experiment seed to 0 for both models.
Success rates weight tasks equally; suite and condition averages use the
groups specified in each result table. Training and policy serving run
separately from the LIBERO simulator, with observations and action chunks
exchanged between the two processes. No simulator rollout is needed for
a ZO training loss evaluation.

\subsection{Hyperparameters and Learning-Rate Sweeps}
\label{app:hyperparameters}

\paragraph{Adaptation methods and execution ablations.}
We compare the unadapted policy with action-side ZO adaptation and
ZO-SGD implementations of FLA~\citep{Li_2026_CVPR} and
DART~\citep{10.1007/978-3-032-37574-2_24}. The daggered baselines replace
the original first-order optimization with two-sided MeZO estimates
followed by SGD. FLA adapts vision-encoder feed-forward projections;
DART adapts source- and target-domain full-model weights separately and
combines their updates using domain arithmetic. Both change the prefix
and therefore cannot use our frozen-prefix cache.
At fixed $q$, the execution ablation holds the adaptation data,
initialization, learning rate, parameter placement, and update count
fixed, progressively adding within-step prefix reuse, persistent reuse
across steps, and schedule-aware prefetch to uncached action-side ZO.

\paragraph{Fixed optimization settings.}
Tables~\ref{tab:adaptation-settings} and~\ref{tab:adaptation-settings-oft}
summarize the $\pi_{0.5}$ and OpenVLA-OFT settings.
All listed stages use task combination 1, seed 0, batch and microbatch size
64, Gaussian two-sided perturbations, and $\epsilon=10^{-3}$. The ZO
update averages $q$ directional estimates and applies SGD without
momentum, weight decay, or gradient clipping. LoRA uses rank and scaling
16, exact-SVD PiSSA initialization~\citep{meng2024pissa}, and trains both
factors.

\paragraph{Learning rate sweep.}
Every OpenVLA-OFT sweep, including the FLA-style and DART baselines, uses
the half-decade learning-rate grid
$\{10^{-5},3\!\times\!10^{-5},10^{-4},3\!\times\!10^{-4},10^{-3},
3\!\times\!10^{-3},10^{-2},3\!\times\!10^{-2},10^{-1}\}$.
Each completed candidate receives a pilot evaluation on the first ten
initial states of every task, or 400 episodes per view.

\begin{table}[t]
\centering\small
\caption{Selected $\pi_{0.5}$ adaptation configurations. $\dagger$ denotes our ZO-SGD baseline implementations. DART's two stages
each start from the same pretrained weights. Execution ablations reuse the
action-side settings at the corresponding $q$.}
\label{tab:adaptation-settings}
\setlength{\tabcolsep}{4pt}
\begin{tabular}{lp{0.33\linewidth}crl}
\toprule
Method & Trainable placement & $q$ & Learning rate & Updates \\
\midrule
Action-side ZO & Action-expert query/value LoRA & 16 & $3\!\times\!10^{-2}$ & 1{,}000 \\
Action-side ZO & Action-expert query/value LoRA & 64 & $10^{-1}$ & 1{,}000 \\
FLA$^\dagger$ & Vision encoder FFN LoRA & 16 & $10^{-3}$ & 1{,}000 \\
DART$^\dagger$ & Full model, source and target & 16 & $10^{-3}$ & $500+500$ \\
\bottomrule
\end{tabular}
\end{table}

\begin{table}[t]
\centering\small
\caption{Selected OpenVLA-OFT adaptation configurations. Action-side LoRA
targets the three eligible linears of the L1 action head, and FLA targets
the SigLIP FFNs of the fused vision backbone. DART trains all VLA,
action-head, and proprio-projector parameters; its two stages each start
from the same pretrained weights. Compound-shift runs reuse these settings
with matching perturbed demonstrations.}
\label{tab:adaptation-settings-oft}
\setlength{\tabcolsep}{4pt}
\begin{tabular}{lp{0.33\linewidth}crl}
\toprule
Method & Trainable placement & $q$ & Learning rate & Updates \\
\midrule
Action-side ZO & Action-head LoRA & 4 & $10^{-2}$ & 1{,}000 \\
Action-side ZO & Action-head LoRA & 16 & $3\!\times\!10^{-3}$ & 1{,}000 \\
Action-side ZO & Action-head LoRA & 64 & $10^{-2}$ & 1{,}000 \\
FLA$^\dagger$ & Vision encoder FFN LoRA & 4 & $3\!\times\!10^{-3}$ & 1{,}000 \\
DART$^\dagger$ & Full model, source and target & 4 & $3\!\times\!10^{-5}$ & $500+500$ \\
\bottomrule
\end{tabular}
\end{table}

\subsection{Timing and Software}
\label{app:timing}

The recorded profiling environment uses NVIDIA RTX PRO 6000 Blackwell
Server Edition GPUs with 96 GB memory, with one GPU visible to each
training process. 

DART uses two 500-update training windows and a separately timed CPU
merge. The merge timer covers checkpoint validation and loading, domain
arithmetic, output serialization, and hashing. Full adaptation accounts
for source training, target training, and this merge; when a source
checkpoint is reused across target conditions, target-only adaptation
excludes the shared source stage.

\section{Implementation of Frozen-State Reuse and Prefetching}
\label{app:prefetch}
\label{app:exactness}

This appendix specifies the cache boundary, minibatch schedule, and transfer protocol behind Section~\ref{sec:cache-prefetch}.

\subsection{Cache Contents}

\paragraph{The reuse boundary of $\pi_{0.5}$.}
For $\pi_{0.5}$, a cached conditioning state contains the attention keys and values from every frozen vision--language layer, together with the prefix padding mask.
For a single observation, each layer's key and value tensors have shape $[1,H_{\mathrm{KV}},L,d_h]$, and the Boolean mask has shape $[1,L]$.
Here, $L$ is the padded prefix length, $H_{\mathrm{KV}}$ the number of key/value heads, and $d_h$ the head dimension.
This state contains the information required by the action expert from the frozen prefix.
Action targets, sampled action noise, flow time, and trainable action projections remain outside the cache.

\paragraph{The reuse boundary of OpenVLA-OFT.}
For OpenVLA-OFT, a cached conditioning state contains the final-layer hidden states of the frozen language model at the action-token positions.
For a single observation, this tensor has shape $[1,KD_a,d]$, and the L1 head reshapes it to $[1,K,D_a d]$ before its first projection.
Here, $K=8$ is the action-chunk length, $D_a=7$ the action dimension, and $d=4096$ the language-model width.
Both camera images, the proprioceptive state, and the prompt are processed before this boundary by the frozen vision backbone, projectors, and language model, and the action head reads no other input; the state therefore contains all information the head requires from the frozen VLA.

\paragraph{Cold and mixed minibatches.}
Cache construction is lazy and takes place in the training thread.
If every sample in a microbatch is cached, its states are loaded in sampler order and prefix computation is skipped.
If any record is absent, the implementation executes the frozen prefix at the original full microbatch shape.
It writes only new records and checks that previously stored rows match the recomputed values exactly.
Keeping the original shape avoids changing matrix-multiplication and attention kernels by compacting only the missing samples.
Consequently, a mixed microbatch can recompute cached observations during cache population: there is at most one stored record per observation and namespace, but there can be more than one physical prefix evaluation of that observation.
The $N c_{\mathrm{pre}}$ population term in Section~\ref{sec:across-step-reuse} describes ideal per-observation materialization; this conservative fallback can add cold-cache computation.
Missing files fall back to demand-time computation, whereas corrupt records, incompatible layouts, changed inputs, or detected numerical mismatches raise an error.

\subsection{Reconstructing the Next Minibatch}

Let $b$ be the microbatch size and $a$ the number of accumulated microbatches, so that the effective batch size is $B=ab$.
For update $t$, the training thread first obtains all $a$ microbatches and their cached or freshly computed prefix states.
It then schedules the $a$ microbatches of update $t+1$ before evaluating the current update's $2q$ probes.
The queue is bounded to one future optimizer update.
Only fully cached future microbatches are eligible; incomplete ones remain on the ordinary demand path.
A future that is not ready when requested delays its own batch and never changes the sample order.
The implementation requires standard ordered batching and the configured full microbatch shape.
The consumed cursor, optimizer state, and training random-number state are checkpointed; speculative futures are rebuilt after resume, and persistent files can either be reused or repopulated lazily.

\subsection{Host Staging and CUDA Streams}

The cache persists on disk, while a single background worker reads records and assembles a complete microbatch in reusable pinned host buffers.
On a dedicated CUDA stream, the worker allocates device tensors with matching shapes, dtypes, and strides, enqueues nonblocking host-to-device copies, and records a completion event.
Before consuming a staged batch, the compute stream waits on that event and records its use of the device tensors through \texttt{record\_stream}.
This establishes the copy-to-compute dependency without synchronizing the entire device.
The host buffers remain alive and unchanged until their copy event completes; only then can they return to the staging pool.
The worker performs file reads, tensor assembly, and transfers only.
All model calls, cache-miss computation, and objective randomness remain in the training thread.

\subsection{What Reuse Costs and What It Preserves}
\label{sec:costs-and-semantics}

\paragraph{What reuse costs.}
The price of all three mechanisms is memory, and it is bounded by the demonstration rather than by the workload.
The accelerator staging budget covers two batches of conditioning states: the $B$ the current update uses, and the $B$ the prefetcher is staging for the next one.
Their payload is therefore at most $2BS$, where $S$ is the size of one conditioning state $h_i$ and depends on how the interface between the frozen prefix and the action expert is represented in each architecture. This excludes the action expert's working tensors and cloned attention cache (Appendix~\ref{app:prefetch}).
For $\pi_{0.5}$, $h_i$ consists of the prefix attention key--value states from 18 VLM layers and occupies approximately $17.02$\,MiB per sample in BF16.
In contrast, OpenVLA-OFT caches a $56\times4096$ action-token hidden state, requiring only $0.4375$\,MiB per sample in BF16.
On the host, the store grows to $N$ records and no further, regardless of $q$ or $T$.

\paragraph{What reuse preserves.}
\label{sec:reuse-preservation}
\system{} changes only the execution of ZO adaptation: it does not alter the objective, perturbation
directions, trainable parameters, or update rule.
For $\pi_{0.5}$, the cached conditioning state is computed from the frozen vision--language prefix and
fixed inputs, so reusing it is neutral to the finite-difference estimator in exact arithmetic.
Action noise and flow time remain sampled for each direction and are replayed within its $\pm$ pair.
Bitwise-identical training trajectories are not implied: the cached path separates prefix prefill from
action denoising and uses a different GPU execution path, while deterministic CUDA algorithms are not
enforced.
Small numerical differences in the two losses can be amplified over
successive ZO updates.
Empirically, our cache-mode ablation in Table~\ref{tab:main} finds success-rate differences comparable to those between repeated executions of an identical configuration.

\end{document}